\documentclass[a4paper,fleqn]{cas-dc}

\usepackage{graphicx}
\usepackage[numbers]{natbib}
\usepackage{hyperref}
\usepackage{booktabs}
\usepackage{caption}
\usepackage{subfigure}
\usepackage{color}

\usepackage{enumitem}

\def\tsc#1{\csdef{#1}{\textsc{\lowercase{#1}}\xspace}}
\tsc{WGM}
\tsc{QE}

\usepackage{booktabs}
\usepackage{array, caption, threeparttable}
\usepackage[font=small,labelfont=bf,labelsep=none]{caption}

\begin{document}
\let\printorcid\relax
\let\WriteBookmarks\relax
\def\floatpagepagefraction{1}
\def\textpagefraction{.001}

\shorttitle{MSCViT: A Small-size ViT architecture with Multi-Scale Self-Attention Mechanism for Tiny Datasets}    

\shortauthors{Y.Zhang,B}  

\title [mode = title]{MSCViT: A Small-size ViT architecture with Multi-Scale Self-Attention Mechanism for Tiny Datasets}  



%





\author[1]{\textcolor{black}{Bowei Zhang}}
\author[1]{\textcolor{black}{Yi Zhang}}
\cormark[1]
\cortext[1]{Corresponding author.}

\affiliation[1]{organization={Department of Computer Science},
                addressline={Sichuan University}, 
                country={China}}

\begin{abstract}
Vision Transformer (ViT) has demonstrated significant potential in various vision tasks due to its strong ability in modelling long-range dependencies. However, such success is largely fueled by training on massive samples. In real applications, the large-scale datasets are not always available, and ViT performs worse than Convolutional Neural Networks (CNNs) if it is only trained on small scale dataset (called tiny dataset), since it requires large amount of training data to ensure its representational capacity. In this paper, a small-size ViT architecture with multi-scale self-attention mechanism and convolution blocks is presented (dubbed MSCViT) to model different scales of attention at each layer. \textcolor{red}{Firstly, we introduced wavelet convolution, which selectively combines the high-frequency components obtained by frequency division with our convolution channel to extract local features.} Then, a lightweight multi-head attention module is developed to reduce the number of tokens and computational costs. Finally, the positional encoding (PE) in the backbone is replaced by a local feature extraction module. Compared with the original ViT, it is parameter-efficient and is particularly suitable for tiny datasets. Extensive experiments have been conducted on tiny datasets, in which our model achieves an accuracy of 84.68\% on CIFAR-100 with 14.0M parameters and 2.5 GFLOPs, without pre-training on large datasets. 
\end{abstract}



\maketitle


\section{Introduction}
CNNs dominated computer vision field in early years. Recently, Transformer begin to prevail, especially ViT \cite{dosovitskiy2020image} excels in modelling and capturing long-range dependency between tokens. However, without sufficient training data, earlier attention layers have limited learning abilities for local information. For instance, the original ViT was pre-trained on large scale dataset JFT-300M and was fine-tuned on ImageNet-1K. However, if it was only trained on ImageNet-1K, it is inferior to CNNs. In depth exploration of the above phenomenon reveals the following reasons:

Firstly, the original ViT lacks inductive bias. Although Pre-training on large-scale datasets compensates for this deficiency, helping ViT to learn stronger representations. In real applications, large-scale datasets are not always accessible, while pre-training followed by fine-tuning is also less desirable and unattainable. In this light, most of the previous methods modified ViT into hierarchical structures and combined convolutional computations within these structures. As a result, such hybrid structures behave more like CNNs, which are more competitive on large-sized datasets (e.g. ImageNet-1K), but there is still a performance gap on tiny datasets.

Secondly, CNN has a natural advantage in learning local features.When dealing with sparse training data, the spatial correlation in the data is often insufficient, while CNN can capture local features, learn local information at lower levels, and integrate these local information at higher levels to obtain global information. On the contrary, according to \cite{raghu2021vision}, without sufficient data, ViT has poor learning abilities for local information in earlier layers. Although ViT has a unified representation and contains more global information in different layers, it exhibits higher similarity across different layers, that means ViT aggregates more global attention in earlier self-attention layers, ignoring local attention (i.e. without sufficient training data, ViT would not learn to attend locally in earlier layers).

In this paper, we aim to create a hybrid model that outperforms both CNN and Transformer on tiny dataset. During the attention computation stage, we extract fine-grained features and integrate coarsegrained features by stacking different-sized convolutional blocks in different attention heads. Additionally, we select deep convolution to merge attention tokens through experiments. Furthermore, to further enhance feature representation, we partially convolve input tokens, selecting redundant attention channels for local convolution calculation. This enables the fusion of local information extracted by convolution when interacting between different attention channels. Moreover, the traditional PE has been replaced by a local information encoding block, which not only provides inductive bias in an implicit way for the entire network structure but is also more beneficial to training on tiny datasets.

\begin{figure}[tbp]\rmfamily
    \centering
    \includegraphics[width=\linewidth]{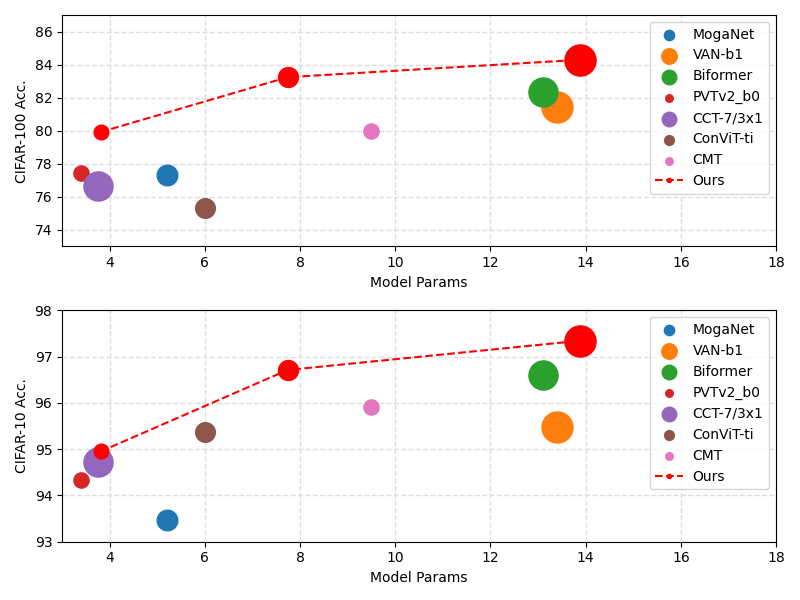}
    \caption{\rmfamily{\textbf{Performance  of MSCViT on CIFAR-10 and CIFAR-100. }MSCViT performs better than some models with similar structures.}}
    \label{fig:1}
\end{figure}

Our model has been tested on popular tiny datasets, including CIFAR-10 \cite{krizhevsky2009learning} and CIFAR-100 \cite{krizhevsky2009learning}. We also select datasets with extremely imbalanced training samples per class, such as Flowers102 \cite{nilsback2008automated}, and datasets with fewer classes, such as Chaoyang \cite{zhu2021hard}. Our results on each dataset are trained from scratch rather than transferring the pre-trained model. We develop 3 models of different scales, namely -tiny, -xs, and -s(small). As shown in Fig. \ref{fig:1}, on CIFAR-100, our xs model achieves top-1 accuracy of 83.44\% with 7.8M parameters and 1.0 GFLOPs of computation. Meanwhile, our small model achieves 84.68\% top-1 accuracy with 14.0M parameters and 2.5 GFLOPs of computation. Compared with other models with similar scales, our model is competitive without pre-training on large datasets. In summary, the contribution of this paper can be summarized as follows:

\begin{itemize} [leftmargin=*]
\item A Local Feature Extraction (LFE) module has been proposed, which is used to capture local information of the intermediate features and replace the PE in the original ViT network.
\item  We study the role of deep convolution in lightweight token merging for attention computation. Based on this, a Lightweight Multi-scale Self-Attention (LMSSA) has been proposed to improve the model’s fine-grained feature extraction ability.
\item A Convolutional Feature Fusion (CFF) module has been presented to enhance the local feature extraction capability of input channels, which obtains object shape by \textcolor{red}{using the combination of wavelet and conventional convolutions.} It adjusts the relationship between convolution locality and attention globality between different channels in the channel dimension, improving the performance of hybrid structures on tiny datasets. 
\end{itemize}

\section{Related works}
\subsection{Vision Transformer (ViT)}

\textcolor[RGB]{46,139,87}{Albeit success in NLP, the self-attention mechanism in the original ViT often overlooks the detailed local features.} To address this issue, DeiT \cite{touvron2021training} used distillation tokens to transfer CNN-based features into ViT. T2TViT \cite{yuan2021tokens} introduced tokenization modules to consider neighboring pixels, recursively rearranging images into tokens. DETR \cite{zhu2020deformable,carion2020end} input locally extracted features from CNNs into a Transformer encoder to model the global relationships between features in a serial way. CrossViT \cite{chen2021crossvit} processed patches of different sizes using a dual-branch Transformer, while LocalViT \cite{li2021localvit} integrated deep convolutions into ViT to improve the local continuity of features. \textcolor[RGB]{46,139,87}{The above works discussed the problem of insufficient feature learning, and proposed different solutions to it. Encouraged by them, we also focus on enhancing local feature extraction ability of ViT on tiny datasets. }

To accomplish dense prediction tasks such as object detection and semantic segmentation, some methods \cite{wang2021pyramid, liu2021swin, zhu2023biformer, wu2021cvt} introduced pyramid structures from CNNs to ViT backbones. PVT \cite{wang2021pyramid} introduced a pyramid structure into ViT, generating multi-scale feature maps for various pixel-level dense prediction tasks. Swin Transformer \cite{liu2021swin} replaced fixed-size positional embeddings with relative positional biases and limits self-attention within shifting windows. Twins \cite{chu2021twins} combined local and global attention mechanisms to obtain stronger feature representations. \textcolor[RGB]{46,139,87}{As can be seen, multi-scale feature extraction play critical role in dense prediction tasks. }

In order to improve the prediction accuracy of the attention mechanism in ViT, some works focus on improving the attention mechanism itself. For example, MaxViT \cite{tu2022maxvit} used blocked local attention and expanded global attention to compose a multi-axis attention mechanism, enabling global and local spatial interactions for arbitrary inputs. Biformer \cite{zhu2023biformer} achieved more flexible computation allocation and content-aware dynamic sparse attention by proposing a novel two-level routing attention mechanism, thereby improving computational efficiency and performance.

Other works focused on improving the attention mechanism in ViT. MaxViT \cite{tu2022maxvit} used blocked local attention and expanded global attention to compose a multi-axis attention mechanism. Biformer \cite{zhu2023biformer} proposed a two-level routing attention mechanism to realize flexible computation allocation. \textcolor[RGB]{46,139,87}{Unlike them, we optimize attention mechanism in different way, where we developed a lightweight multi-scale self-attention module to reduce the number of tokens and computational costs, while enhancing feature extraction capabilities at different granularities. }

\begin{figure*}[htbp]\rmfamily
  \centering
  \includegraphics[width=\textwidth]{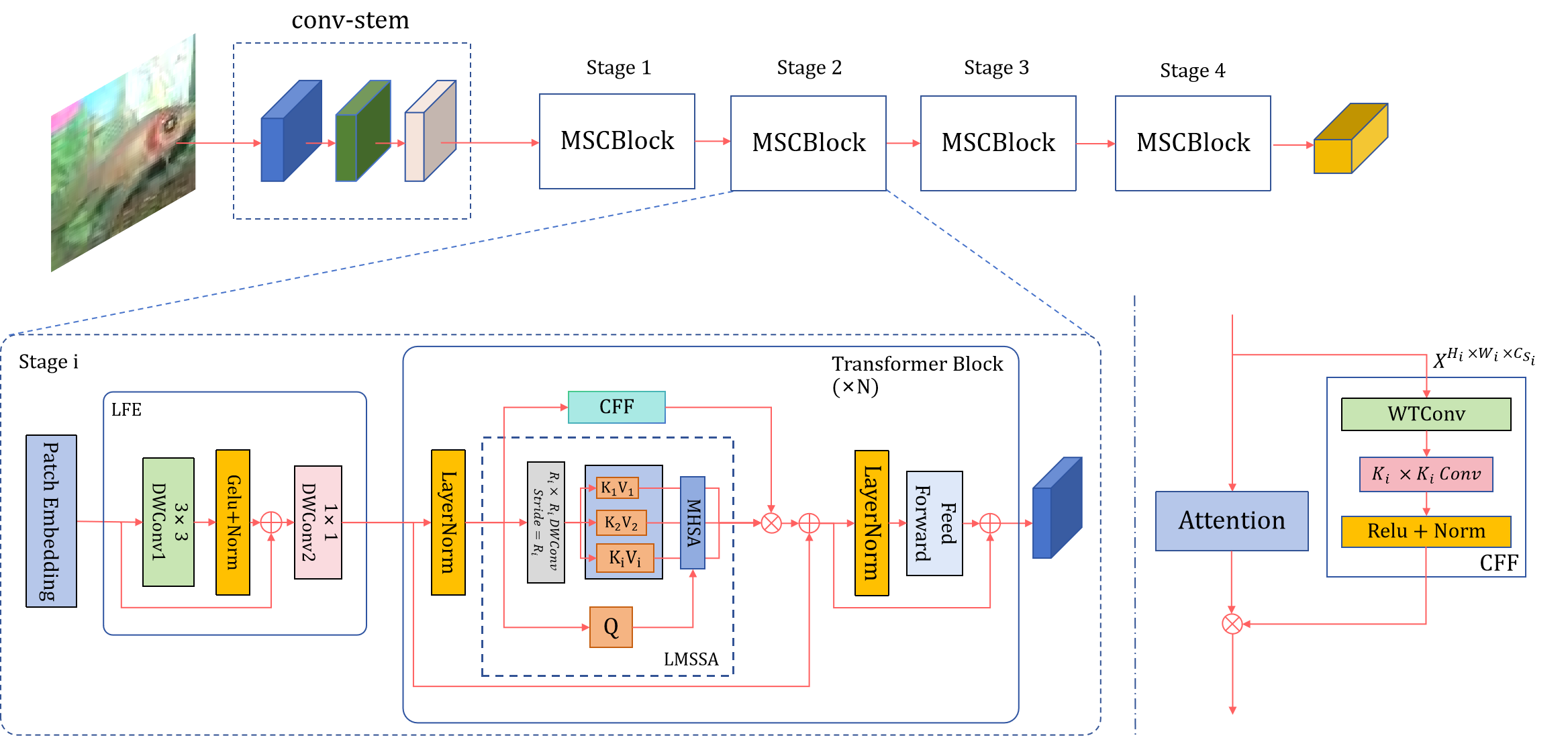}
  \caption{\rmfamily{The overall architecture of the proposed MSCViT.}}
  \label{Fig1}
\end{figure*}

\subsection{Introducing Convolutions to Transformer}

Due to the lack of inductive bias, pure Transformer based visual models have poor generalization ability. \textcolor[RGB]{46,139,87}{A possible solution is to combine the attention layers with convolution layers.} CoAtNet \cite{dai2021coatnet} stacked convolutional layers and attention layers in the model architecture. CCT \cite{hassani2021escaping} adopted convolution tokenization modules and replaced class tokens with a final sequence pooling operation. The Feature Coupling Units (FCUs) in Conformer \cite{peng2021conformer} interactively fuses local features and global representations at different resolutions. CvT \cite{wu2021cvt} introduced convolutions by embedding a new convolution token and using convolutional projection in a convolution Transformer block. \textcolor[RGB]{46,139,87}{We notice that the above works simply integrated CNN and Transformer without differentiating the channels. Considering this, we chose deep convolution to merge the attention tokens while partially convolving the input channels to fuse the feature information extracted from the input blocks. This will help narrow the performance gap of ViT on tiny datasets. }

\subsection{ViT for tiny datasets}
Despite the favorable results on large-scale dataset (e.g. ImageNet-1K), most of the existing models cannot compete with CNNs if they are only trained on tiny datasets (e.g. CIFAR-100). \textcolor[RGB]{46,139,87}{To bridge the performance gap}, Liu, Yahui, et al. \cite{liu2021efficient} introduced self-supervised style training strategies and loss functions to carry out training on tiny datasets. SL-ViT \cite{lee2021vision} utilized shifted patch tokenization modules and modified self-attention to make the model focus more locally. TransMCGC \cite{xiang2023transmcgc} enhanced feature learning capabilities by replacing inefficient Transformer blocks in the final stage with convolutional MCGC blocks. SATA \cite{chen2023accumulated} reduced attention noise by separating trivial attention weights and adjusting them as part of the maximum attention weights. \textcolor[RGB]{46,139,87}{Unfortunately, the above methods overlooked the advantages of combining convolutional features with attention features. The combination of the 2 features enables the model to better distinguish different details and their relationships in the image (even under scarce data), which also makes the model focus on the most relevant features while ignoring noise and irrelevant details in tiny datasets.} In view of this, our proposed approach leverages the characteristics of convolutions to reconcile the relationship between convolutional locality and global dependency so as to obtain the performance gain on tiny datasets.

\section{Method}
\subsection{Overall Architecture}
In this paper, we aim to build a small-size vision Transformer network for tiny datasets by combining the advantages of CNNs and ViT. The overall architecture of our proposed network is shown in Fig. \ref{Fig1}. It is reported \cite{xiao2021early} that compared with direct partition and downsample operation, using convolution to downsample the input images at the beginning of the network can better extract local information. Based on this, we construct a Conv-stem by utilizing one 3×3 convolution for downsampling along with two convolution blocks with kernel size 3, stride 1 and padding 1. After each convolution, we apply Batch Norm and GELU to stabilize the network and to improve model’s generalization ability. Meanwhile, in each stage, we stack different numbers of transformer blocks to construct models of different sizes. \textcolor[RGB]{46,139,87}{The value of N in Fig. \ref{Fig1} for different stages of MSCBlock corresponds to the Depth values in Table \ref{Table1} (for different versions of MSCViT).} Each transformer block consists of the same structure, including LFE, LMSSA, CFF and FFN (which will be discussed in section 3.2 and section 3.3 in details). Finally, the entire model is ended by a global average pooling layer and a classification layer with softmax, which yields dense prediction outputs.

\subsection{Local Feature Extraction (LFE)}
The original ViT divides the input image into different tokens for attention computation. Since the self-attention mechanism is permutation-invariant, the input sequence has no inherent order. Therefore, the original ViT introduces PE to realize sequence awareness, which can be either learnable or fixed. 

Studies \cite{dosovitskiy2020image} have shown that removing PE causes significant feature loss. To mitigate the loss, we introduce small-sized convolutional blocks to implicitly encode positional information. Leveraging the translation invariant feature of convolutions, our proposed LFE can  effectively utilize data augmentation techniques such as rotation and translation without compromising model performance caused by removing PE. Additionally, adding small-sized convolutions in multiple stages of the overall architecture effectively extracts local information from feature maps, mitigating ViT's deficiency in capturing the local structural information of tokens. Since tiny datasets often have small-scale characteristics, this approach is effective for applying the model to small-scale datasets.

The calculation process of our LFE can be expressed as follows:\textcolor[RGB]{46,139,87}{{\small
\begin{equation}\label{eq1}
\mathrm{Local(X)=DWConv2(GELU(BN(DWConv1(X)))+X)   }
\end{equation}
}}Here the input $\mathrm{X \in R^{H \times W \times C}}$, $\mathrm{H \times W}$ represents the size of the input features and $C$ denotes the current feature channels. DWConv1 is a 3×3 depth-wise convolution block. \textcolor{blue}{It is used to extract detailed local features from the input feature map, which is crucial for enhancing the model’s understanding of fine-grained structures in the data.} DWConv2 is a 1×1 depth-wise convolution block, \textcolor{blue}{which is used to adjust the dimensionality of the channel, balancing computational complexity without affecting accuracy.} Experimental results indicate that during training on tiny datasets, our proposed LFE provides inductive bias capability, which avoids the performance degradation after the removal of the PE in the original ViT.

\subsection{Lightweight Multi-scale Self-Attention (LMSSA)}
For the attention module, each input feature map X is projected into Q, K and V, and then attention results are calculated in parallel through N independent attention heads. However, the high computational cost of calculating the original attention makes the training of attention inefficient. Therefore, we further utilize multi-scale information fusion on the basis of token feature fusion. Specifically, for the original feature fusion, the index of each attention calculation head is $i$, then it is written as:{
\textcolor[RGB]{46,139,87}{
\begin{align}
\mathrm{Q}_{i} &=\mathrm{X}           W_{ q_{i} } \\
\mathrm{K}_{i} &=\mathrm{Reshape(X,R)}W_{ k_{i} } \\
\mathrm{V}_{i} &=\mathrm{Reshape(X,R)}W_{ v_{i} }
\end{align}}}\textcolor{blue}{Here $Reshape$ denotes token feature fusion, R is the fusion coefficient. Specifically, Reshape stands for reshaping operation, i.e. turning the dimension of the feature map X into $\mathrm{\frac{HW}{R^{2}} R^{2}C } $ , where $\mathrm{HWC}$ is the input feature of the current layer (with size Height × Width × Channel).} For single-scale fusion, R is the same for each attention head. During the computation of multi-head, all heads can be evenly divided into n parts (i.e., head\_0, head\_1, ... head\_n). Different fusion coefficients $R_{i}$ are set for each attention head. In our backbone, we set $R_{i}$ to {8, 4, 2, 1}.Formally:
\textcolor[RGB]{46,139,87}{
\begin{align}
    \mathrm{K}_{i}&=\mathrm{Reshape}(X_{i} , R_{i}) W_{ k_{i} }  \; \; {i\in{1,2,3,4}}\\ 
    \mathrm{V}_{i}&=\mathrm{Reshape}(X_{i} , R_{i}) W_{ v_{i} }  \; \; {i\in{1,2,3,4}}
\end{align}}Here, because different feature channels use different fusion coefficients, Therefore, $X_{i}$ selects different feature channels for each K and V. The, the calculation of a single attention head is expressed as follows:\textcolor[RGB]{46,139,87}{
\begin{equation}
\mathrm{Attention} \;i( q^{i},k^{i},v^{i} ) = \mathrm{Softmax}( \frac{ q^{i}  {k^{i}}^{\mathrm{T}} }{ \sqrt{ d_{k} } } ) v^{i}
\end{equation}}After the computation of each attention head, they are concatenated. Then, to reduce the computational cost and improve the inference speed, we use depth-wise convolution block with a step size of $k \times k$ for feature fusion, which reduces the spatial size of K and V before the attention operations as $\mathrm{K'} = \mathrm{DWConv(K)}, \,  \mathrm{K} \in \mathrm{R}^{ \frac{n}{ k^{2} }  \times  d_{k}  }$ and $\mathrm{V'} = \mathrm{DWConv(V)}, \,  \mathrm{V} \in \mathrm{R}^{ \frac{n}{ k^{2} }  \times  d_{v}  }$. \textcolor{brown}{Traditional self-attention mechanisms (e.g. SRA), operate at a single scale, which may not be sufficient to capture the various information exists in the datasets. By contrast, the head in our lightweight attention with large fusion coefficient focuses on extracting coarse-grained features, which depict the overall structure and background of the image and quickly identify the main objects and estimate their positions. As the network deepens, the head with smaller fusion coefficient will be responsible for extracting finer-grained details, which are crucial for identifying specific parts of an object.} Ablation experiments in subsequent section show that performing lightweight multi-scale attention computation does not sacrifice too much accuracy, therefore our proposed lightweight structure is reasonable and effective. 

\subsection{Convolutional Feature Fusion (CFF)}

The original ViT's attention mechanism excels in long-range modeling, which achieves good results with abundant training data. However, due to its lack of capability in extracting local features, its performance on tiny datasets is quite limited. We integrate convolutions into the attention computation process so as to reap the benefit of both of them. Meanwhile, research \cite{liu2023efficientvit} has shown that some channels in attention computation are inefficient and redundant. 

\textcolor{blue}{In this light, we select specific channels for convolutional feature fusion, then wavelet convolution is adopted to extract the high-frequency features of the image so as to learn shape information of the object while expanding large receptive field. Formally:}\textcolor[RGB]{46,139,87}{
\begin{align}
\mathrm{Conv}_{St} &= \mathrm{Conv(WTConv(X,} S_{i}),\mathrm{K} = ki)  \\
\mathrm{X’} &= \mathrm{GELU(Norm(Conv} _{st} (\mathrm{X}, S_{i} )))
\end{align}}\textcolor{blue}{Here,WTConv represents wavelet convolution operation.  $S_{i}$ refers to the weight for the input feature map X, in which the corresponding number of channels are selected (based on the layers of the stage) to participate in the operation of CFF. This operation enable the model to measure the importance of features and to balance the computational cost and performance. $\mathrm{Conv}_{st}$ represents the convolution operation for different stages.} Experiments indicate that using different kernel sizes for different levels is beneficial to feature extraction.

On the other hand, considering that ViT captures more local information in shallow layers while it captures global features in deeper layers, we use convolutional kernels of different sizes and paddings at different levels of the backbone network to adapt to this situation. For the channels without convolutional feature fusion, we apply the original attention computation method. Finally, the feature map output by the attention mechanism is: \textcolor[RGB]{46,139,87}{
\begin{equation}
\mathrm{X_{out} = concat(X’,Attention(X}_{ S_{i}' } ))
\end{equation}}The original ViT architecture (e.g. PVT) consists of an attention module and a feed-forward module (FFN), which is responsible for nonlinear transformation of the input in the Transformer, thereby enhancing the model's representation ability. The calculation process of FFN is conductive to the utilization of implicit positional information fused by convolutional features. It not only weakens the impact of removing the PE in the backbone, but also improves the utilization efficiency of input features, thereby improving the performance on tiny datasets.

\subsection{Scaling Strategy}

In section \ref{AB:CFF}, we will discuss the optimal size of the convolutional feature fusion module, we will see that selecting different structures for different layers can better improves the experimental results. However, it is different for the LMSSA, since the original PVT uses different scaling factors for each layer. Specifically, the lower layers have larger scales, while the higher layers have smaller scales. Since the input of the last layer has been reduced to 7×7 under standard input size (224×224). On the other hand, it will be more easier to choose the appropriate scaling factor for lower layers (than for higher layers).  For the original ViT, assume the feature dimension of the input is n×d, then the computation cost for the Multi-Head Attention computation is:\textcolor[RGB]{46,139,87}{
\begin{align}
\mathrm{O(MHSA)} &= 4nd^{2} + 2n^{2}d  \\
\mathrm{O(FFN)}  &= 8nd^{2} 
\end{align}
}Let $R_{i}$ be the scaling factor for a certain layer, then the computational workload of this layer after spatial reduction operation will be:
\textcolor[RGB]{46,139,87}{
\begin{equation}
\mathrm{O(LMSSA)} = 4n d^{2} +  \sum_{i}  \frac{2n^{2}d}{  R_{i} ^{2} }
\end{equation}
}Compared with standard Transformer, our model has lower computation cost, which makes it easier to process features at higher resolutions (i.e. larger n).

\begin{table}[tb]\rmfamily
\caption{The number of samples and classes of different tiny datasets.}
\centering
\begin{tabular*}{\linewidth}{@{\extracolsep\fill}cccc@{\extracolsep\fill}}
\toprule
dataset&	train&	test&	class\\
\midrule
CIFAR-10&	50000	&10000&	10\\
CIFAR-100&	50000&	10000&	100\\
Flowers102&2040&	6149&	102\\
Chaoyang&	4021&	2139&	4\\
Oxford-IIIT Pet & 3680&3669 & 37 \\
TinyImageNet&	100000&	10000&	200\\

\bottomrule
\end{tabular*}
\label{Table2}
\end{table}

\section{Experiment}

Extensive experiments have been conducted on tiny datasets, including CIFAR-100 \cite{krizhevsky2009learning}, CIFAR-10 \cite{krizhevsky2009learning}, Flower 102 \cite{nilsback2008automated} and Chaoyang \cite{zhu2021hard}, to validate the effectiveness of our model. In addition, the results on CIFAR-100 is also reported to assess the impact of the core components of our model through ablation.

\subsection{Datasets}
Our goal is to construct a mobile-friendly ViT architecture which is suitable for training from scratch on tiny datasets. The descriptions of the tiny datasets are as follows:
\begin{itemize} [leftmargin=*]
\item CIFAR-10: It consists of 60,000 images with a resolution of 32×32, which are divided into 10 classes with 50,000 training data instances, averaging 5,000 instances per class.
\item CIFAR-100: It consists of 60,000 color images with a resolution of 32x32. The 100 classes in CIFAR-100 are grouped into 20 super-classes. Each image has a "fine" label (the class it belongs to) and a "coarse" label (the super-class it belongs to). Each class has 600 images, including 500 training images and 100 images for testing.
\item Flower 102: This dataset comprises 8,189 images and 102 flower categories, with a total of 2,040 training images and 6,149 testing images. Each category consists of 40 to 258 images, exhibiting significant variations in scale, pose, and lighting. Additionally, there are categories with significant intra-class variation and several very similar categories.
\item Chaoyang: This is a medical image dataset. The training images include 1,111 normal, 842 serrated, 1,404 adenocarcinoma, and 664 adenoma samples. The test images consist of 705 normal samples, 321 serrated samples, 840 adenocarcinoma samples, and 273 adenoma samples. 
\item \textcolor[RGB]{46,139,87}{Oxford-IIIT Pet: The Oxford-IIIT Pet Dataset has 37 categories with roughly 200 images for each class. The images have a large variations in scale, pose and lighting.}
\end{itemize}
The quantities of training and test sets in the dataset are shown in Table \ref{Table2}.

\subsection{Experiment Settings} \label{ExpSetting}
\textcolor{brown}{Firstly, all experiments were conducted under Linux system 5.8.0 and CUDA version 11.6. The runtime environment includes Python 3.9.18, PyTorch 2.1.0, PyTorch Image Models 0.4.12, and OpenMMLab Computer Vision library 1.3.8.} We trained our model from scratch on the training set and reported the top-1 accuracy on the test set. We used the same data augmentation methods as DeiT without additional hyper-parameter settings. We trained our network via random initialization using the AdamW optimizer with a cosine decay learning rate scheduler. We followed the training settings of DeiT, where we trained 3 different sized models 300 epochs on 2 NVIDIA RTX 3090 GPUs, and set the batch size to 128, the initial learning rate to \textcolor{blue}{5 × $10^{-4}$}, weight decay to 0.05, warm-up epochs to 5. Additionally, our models were implemented using the PyTorch framework instead of TensorFlow.

We name our model MSCViT-S, which has similar model size and computational complexity to DeiT-S and PVT-S. We also develop models of different scales, including MSCViT-T and MSCViT-XS. All models share the same input resolution. The detailed hyper-parameters are shown in Table \ref{Table1}.

\begin{table*}\rmfamily
\centering
\caption{The parameters of our proposed models. The output size corresponds to an input resolution of 224×224. For the Patch Embedding layer, the down-sampling size is shown in parentheses. For Stage 1 - 4, $[C_{k}, P]$ represents the size of the convolutional feature fusion, $R_{i}$ denotes the scaling factor. \textcolor[RGB]{46,139,87}{The value of N in Fig. \ref{Fig1} for different stages of MSCBlock corresponds to the Depth values below}, the number of parameters and computation cost are all listed at the bottom of Table \ref{Table1}.}

\begin{tabular*}{\linewidth}{@{\extracolsep\fill}CCCCC@{\extracolsep\fill}}
\toprule
Input Size&Layer Name&MSCViT-T&MSCViT-XS&MSCViT-S\\
\midrule
224×224 &Conv-stem& 
$\begin{array}{c}
\left[ 3\times3, S=2, P=1,C=16 \right ] \\
{\left[ 3 \times 3, S=1,P=1,C=16 \right] \times 2 }   \\
\end{array} $
& 
$\begin{array}{c}
\left[ 3\times3, S=2, P=1,C=24 \right ] \\
{\left[ 3 \times 3, S=1,P=1,C=24 \right] \times 2 }   \\
\end{array} $
& 
$\begin{array}{c}
\left[ 3\times3, S=2, P=1,C=32 \right ] \\
{\left[ 3 \times 3, S=1,P=1,C=32 \right] \times 2 }   \\
\end{array}$ \\
\midrule
112×112 & Patch Embedding &
$\begin{array}{c}
\left[ 2\times2, S=2,C=32 \right ] \\
\end{array} $
& $\begin{array}{c}
\left[ 2\times2, S=2,C=48 \right ] \\
\end{array} $
& $\begin{array}{c}
\left[ 2\times2, S=2,C=64 \right ] \\
\end{array} $ \\
\midrule
56×56   &Stage1&
$\begin{array}{c}
\left[ C_{k} = 3,P=1 \right ]\\
\left[ R_{1} = 8,R_{2} = 4 \right ]\\
\end{array} $
& $\begin{array}{c}
\left[ C_{k} = 3,P=1 \right ]\\
\left[ R_{1} = 8,R_{2} = 4 \right ]\\
\end{array} $
& $\begin{array}{c}
\left[ C_{k} = 3,P=1 \right ]\\
\left[ R_{1} = 8,R_{2} = 4 \right ]\\
\end{array}$ \\
\midrule
56×56   &Patch Embedding
& $\begin{array}{c}
\left[ 2\times2, S=2,C=64 \right ] \\
\end{array} $
& $\begin{array}{c}
\left[ 2\times2, S=2,C=96 \right ] \\
\end{array} $
& $\begin{array}{c}
\left[ 2\times2, S=2,C=128 \right ] \\
\end{array} $
\\
\midrule
28×28   &Stage2&    
$\begin{array}{c}
\left[ C_{k} = 3,P=1 \right ]\\
\left[ R_{1} = 4,R_{2} = 2,R_{3}=1 \right ]\\
\end{array} $
&    
$\begin{array}{c}
\left[ C_{k} = 3,P=1 \right ]\\
\left[ R_{1} = 4,R_{2} = 2,R_{3}=1 \right ]\\
\end{array} $
&    
$\begin{array}{c}
\left[ C_{k} = 3,P=1 \right ]\\
\left[ R_{1} = 4,R_{2} = 2,R_{3}=1 \right ]\\
\end{array} $
\\
\midrule
28×28   &Patch Embedding
& $\begin{array}{c}
\left[ 2\times2, S=2,C=128 \right ] \\
\end{array} $
& $\begin{array}{c}
\left[ 2\times2, S=2,C=192 \right ] \\
\end{array} $
& $\begin{array}{c}
\left[ 2\times2, S=2,C=256 \right ] \\
\end{array}$ \\
\midrule
14×14   &Stage3& 
$\begin{array}{c}
\left[ C_{k} = 5,P=2 \right ]\\
\left[ R_{1} =2,R_{2} = 1 \right ]\\
\end{array} $
& 
$\begin{array}{c}
\left[ C_{k} = 5,P=2 \right ]\\
\left[ R_{1} =2,R_{2} = 1 \right ]\\
\end{array} $
& 
$\begin{array}{c}
\left[ C_{k} = 5,P=2 \right ]\\
\left[ R_{1} =2,R_{2} = 1 \right ]\\
\end{array} $
\\ \midrule
14×14   &Patch Embedding
& $\begin{array}{c}
\left[ 2\times2, S=2,C=256 \right ] \\
\end{array} $
& $\begin{array}{c}
\left[ 2\times2, S=2,C=384 \right ] \\
\end{array} $
& $\begin{array}{c}
\left[ 2\times2, S=2,C=512 \right ] \\
\end{array}$ \\ \midrule
7×7     &Stage4&
$\begin{array}{c}
\left[ C_{k} = 5,P=2 \right ]\\
\left[ R_{1} =1 \right ]\\
\end{array}$ &
$\begin{array}{c}
\left[ C_{k} = 5,P=2 \right ]\\
\left[ R_{1} =1 \right ]\\
\end{array}$ &
$\begin{array}{c}
\left[ C_{k} = 5,P=2 \right ]\\
\left[ R_{1} =1 \right ]\\
\end{array}$ \\
\bottomrule
\multicolumn{2}{c|}{Depths}  & 
$\begin{array}{c}
\left[ 1,2,4,1 \right ]\\
\end{array}$& 
$\begin{array}{c}
\left[ \textcolor{brown}{1,1,3,2} \right ]\\
\end{array}$& 
$\begin{array}{c}
\left[ 2,2,4,2 \right ]\\
\end{array}$\\
\midrule
\multicolumn{2}{c|}{Params (M)} & 	3.8 &7.8&14.0\\
\midrule
\multicolumn{2}{c|}{GFLOPs}  & 0.5	&1.0&	2.5\\
\bottomrule
\end{tabular*}
\label{Table1}
\end{table*}

\begin{table*}[t]\rmfamily
\centering
\caption{The comparison of classification results on 4 tiny datasets (* in the model annotation represents data from the corresponding paper).}
\begin{tabular*}{\linewidth}{@{\extracolsep\fill}ccccccccc@{\extracolsep\fill}}
\toprule
Models                                                      & \textcolor{blue}{Type}    & Params (M) & GFLOPs & CIFAR100 & CIFAR10 & Flowers102 & Chaoyang  & Oxford-IIIT Pet \\ \hline
PVTv2-b0 \cite{wang2022pvt}                                 & Hybrid & 3.4              & 0.6 & 77.44 & 94.34 & 41.96 & 82.05 & 46.52 \\ 
CCT-7/3x1* \cite{hassani2021escaping}                       & Hybrid & 3.7              & 1.2 & 76.67 & 94.72 & -     & -     &  - \\ 
\textcolor[RGB]{46,139,87}{MogaNet-XT}\cite{li2023moganet}                              & \textcolor{blue}{CNN}     & 3.0 & 1.0 & 74.22 & 93.82 & 40.07 & 79.71 & 42.60 \\
\textcolor[RGB]{46,139,87}{VAN-b0} \cite{guo2023visual}                                 & \textcolor{blue}{CNN}     & 4.1 & 0.9 & 76.10 & 94.37 & 39.53 & 79.42 & 42.70 \\
\textcolor[RGB]{46,139,87}{HSViT-C3A4*} \cite{xu2024hsvit}                              & Hybrid & 2.3              & 1.3 & 72.46 & 93.04 & -     &    -  & -   \\
MSCVIT-T(ours)                                              & Hybrid & 3.8              & 0.5 & \textbf{80.11} & \textbf{95.12} & \textbf{55.86} & \textbf{82.35} & \textbf{53.33} \\ 
\midrule
MogaNet-T \cite{li2023moganet}                              & \textcolor{blue}{CNN}     & 5.2 & 1.1 & 77.33 &93.48	&38.01	&79.85 & 46.55 \\
ConViT-Ti \cite{d2021convit}                                & Hybrid & 6.0              & 1.0 & 75.32 & 95.38 & 57.51 & 82.47 & 30.33 \\ 
CMT-Ti \cite{guo2022cmt}                                    & Hybrid & 9.5              & 0.6 & 79.97 & 95.90 & 56.87 & 79.32 & 56.69 \\ 
HSViT-C4A8* \cite{xu2024hsvit}                              & Hybrid & 6.9              & 1.9 & 73.85 & 94.04 & - & - & - \\
MSCVIT-XS(ours)                                             & Hybrid & 7.8              & 1.0 & \textbf{83.44} & \textbf{96.79} & \textbf{62.35} & \textbf{83.46} & \textbf{60.62} \\ 
\midrule
BiFormer-T \cite{zhu2023biformer}                           & ViT    & 13.1             & 2.2 & 82.32 & 96.61 & 63.95 & 80.36 & 66.83 \\ 
PVT-T \cite{wang2021pyramid}                                & Hybrid & 13.2             & 1.9 & 69.62 & 90.51 & 59.68 & 82.70 & 41.07 \\ 
Swin-T* \cite{liu2021swin}                                  & ViT    & 27.5             & 1.4 & 78.07 & 94.46 & - & -  & -\\ 
LeViT-192 \cite{graham2021levit}                            & Hybrid & 10.9             & 0.6 & 70.24 & 89.22 & 54.48 & 80.97 & 38.15 \\ 
Shunted-T \cite{ren2022shunted}                             & Hybrid & 11.5             & 2.1 & 81.66 & 96.74 & 59.10 & 82.37 & 65.08 \\ 
\textcolor[RGB]{46,139,87}{CMT-XS} \cite{guo2022cmt}                                    & Hybrid & 15.2             & 1.5 & 82.42 & 97.05 & 64.33 & 82.42 & 61.35 \\
CvT-13 \cite{wu2021cvt}                                     & Hybrid & \textcolor{blue}{20.0}             & 4.5 & 81.81 & 89.02 & 54.29 & 81.93 & 60.22 \\ 
van-b1 \cite{guo2023visual}	                                & \textcolor{blue}{CNN}     & 13.9& 2.5 & 81.41 & 95.47 & 42.04 & 81.70 & 57.29 \\
PVT-S \cite{wang2021pyramid}                                & Hybrid & 24.5             & 3.8 & 69.79 & 92.34 & 61.41 & 80.04 & - \\ 
MSCVIT-S(ours)                                              & Hybrid & 14.0             & 2.5 & \textbf{84.68} & \textbf{97.75} & \textbf{65.79} & \textbf{84.11} & \textbf{68.52} \\ 
\bottomrule
\end{tabular*}
\label{Table3}
\end{table*}

\subsection{Results}\label{Res}
\subsubsection{Results on tiny datasets}
The results on tiny datasets are presented in Table \ref{Table3}, where comparison is made on 5 different tiny datasets among our model and mainstream multi-stage ViT and convolution-fused ViT. All methods are tested and compared under the same setting. Firstly, this result proves our earlier claim that previous ViT architectures (e.g. PVT and Swin-Transformer) achieve better results on ImageNet-1K but achieve less satisfactory results on tiny datasets (like CIFAR). Secondly, our model exceeds all other models of similar sizes. To be specific, our model achieves top-1 accuracy of 84.68\% on CIFAR-100 with only 14.0M parameters and 2.5 GFLOPs without pre-training or fine-tuning. Particularly, our tiny version achieves top-1 accuracy of 80.11\% on CIFAR-100 with only 3.8M parameters and 0.5GFLOPs, which is close to or even surpasses some larger-sized models. This indicates that our small-size models perform well on devices with limited resources and computational power.\textcolor{blue}{
It’s also worth noting that CNN is always regarded as the standard for tiny datasets. However, as shown in Table 3, we still lead pure CNN based networks \cite{li2023moganet,guo2023visual,xu2024hsvit} with safe margins. 
}


\subsubsection{Results on Tiny ImageNet}
In this section, experiment has been conducted on Tiny ImageNet to further investigate the performance gap of our models on standard medium-sized datasets and tiny datasets. Tiny ImageNet 200 is a subset of the ImageNet-1K dataset, consisting of 100,000 images, each of them is down-sampled to a size of 64×64 pixels. Tiny ImageNet contains 200 classes, with each class composed of 500 training samples, 50 validation samples, and 50 testing samples. Tiny ImageNet serves as a thumbnail version of ImageNet-1K, with all images down-sampled to 64×64 pixels. However, its training data is far less than that of ImageNet-1K, which places higher demands on the model's feature extraction abilities. Additionally, due to the down-sampling operation, the training of ViT will be more difficult than training CNNs. 

\begin{figure}[tb]\rmfamily
  \centering
  \includegraphics[width=\linewidth]{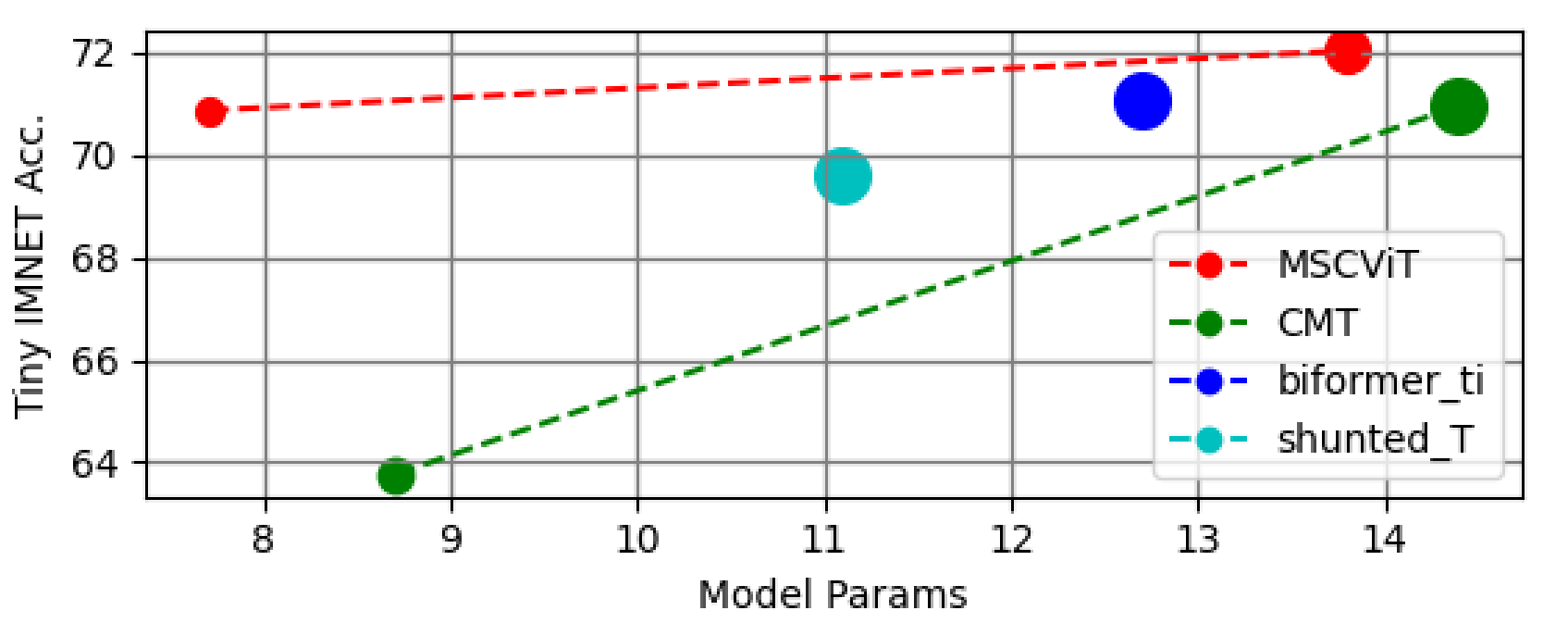}
  \caption{\rmfamily{The comparison of the model sizes and accuracies among different methods.}}
  \label{Fig2}
\end{figure}

In this section, all settings are the same as in section \ref{ExpSetting}. We selected some ViT models with similar structures (including fusion of convolutions) for comparison.  The experimental results are shown in both Table \ref{Table5} and Fig. \ref{Fig2}. Our model achieved 72.11\% accuracy, while shuntedViT and CMT had poor performances. This is because Tiny ImageNet is a subset of ImageNet, whose data volume is much less than ImageNet. Therefore, the traditional ViT model could not learn enough features on this dataset. Moreover, due to the low image resolution, the hybrid models cannot fully make use of their advantages. However, our model can better capture global and local features through multi-scale attention methods, and can also compensate for the shortcomings of fine-grained recognition through convolutional fusion while focusing on the recognition subject.

\begin{figure*}[tb]\rmfamily
  \centering
  \includegraphics[width=\linewidth]{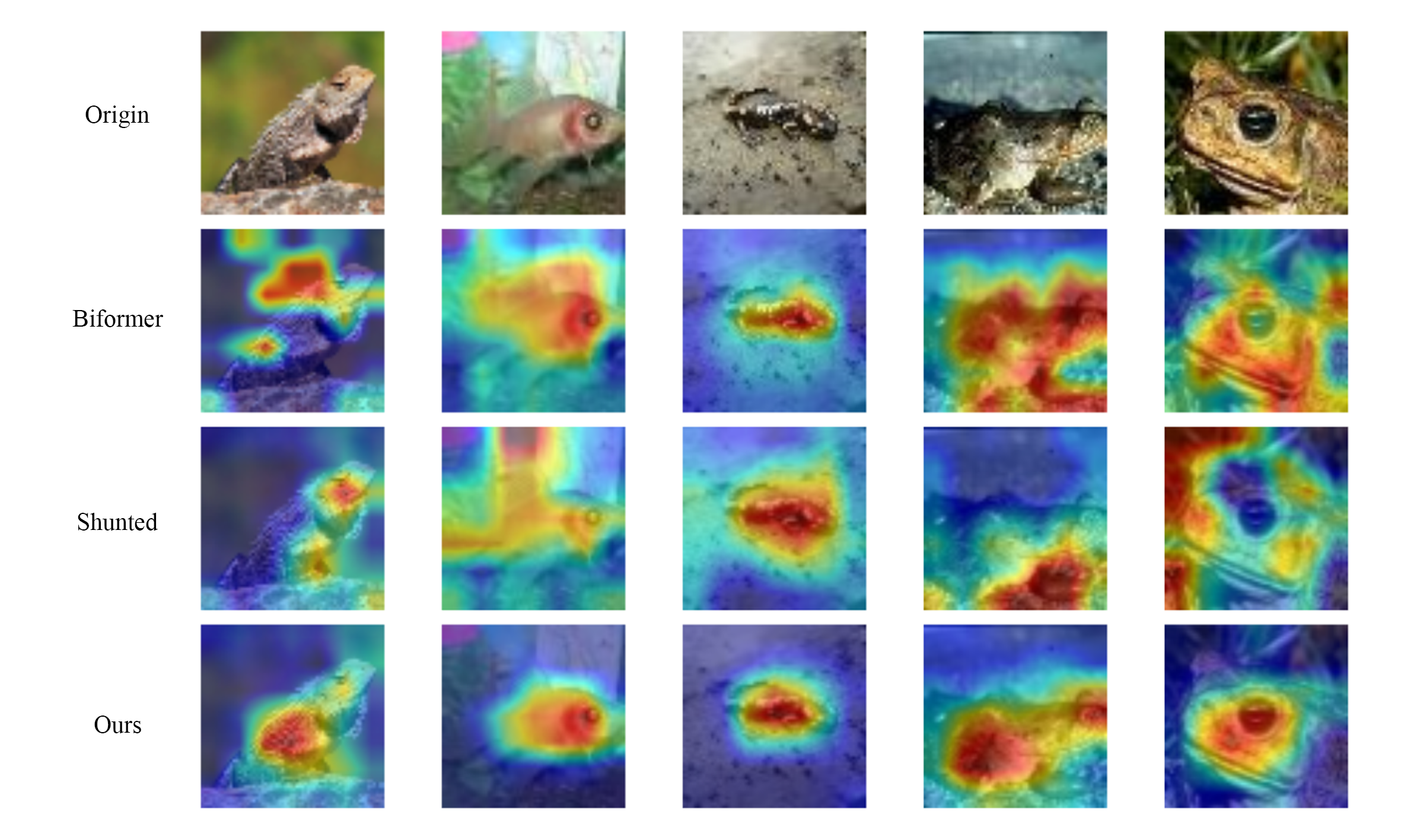}
  \caption{\rmfamily{The comparison  of heatmaps of different methods generated by Grad-CAM.}}
  \label{Fig3}
\end{figure*}

Besides, Grad-CAM is carried out on 5 different categories which reflects the feature heatmaps generated by the final block. As shown in Fig. \ref{Fig3}, BiFormer\cite{zhu2023biformer} and Shunted\cite{ren2022shunted} capture the less important or irrelevant parts, while we captures the attention scattered in the background and focus more on the object itself.

\begin{table}[t]\rmfamily
\caption{The comparison of a comprehensive results Tiny ImageNet.}
\centering
\begin{tabular*}{\linewidth}{@{\extracolsep\fill}ccccc@{\extracolsep\fill}}
\hline
Models& Type & Params  & GFLOPs & Tiny Acc.(\%)  \\ \hline
cmt\_Ti\cite{guo2022cmt}& Hybird & 8.7 & 0.6 & 63.75 \\ 
cmt\_XS\cite{guo2022cmt}& Hybird  & 14.4 & 1.5 & 70.96  \\ 
BiFormer\_T\cite{zhu2023biformer}& ViT  & 12.7 & 2.2 & 71.07 \\ 
Shunted\_T\cite{ren2022shunted}& Hybird  & 11.1 & 2.1 & 69.61  \\ 
MSCVIT-S(ours)& Hybird  & 14.0 & 2.5 & \textbf{72.11}  \\ \hline
\end{tabular*}
\label{Table5}
\end{table}

\subsection{Ablation Study}
In this section, several ablation experiments have been conducted to verify the effectiveness of our proposed model. 
\subsubsection{The overall effect of core components}
In this section, ablation experiments have been conducted on CIFAR-100 to validate the 3 core components of our model, including LFE, LMSSA and CFF. The results are shown in Table \ref{Tableab}, for each combination, the model is trained 200 epochs on a single 3090 GPUs. Without any modules, we achieve the lowest results. With the addition of LMSSA, the accuracy has been increased by 0.86\%, while the computation cost also increases accordingly. With LMSSA+LFE or LMSSA+CFF, we enjoy slight increase in accuracies. Since CFF module performs convolutional feature extraction by utilizing certain attention channels, the computation cost actually decreases. LFE module is mainly used to replace PE, makes a negligible impact on the computation cost. Finally, with LFE+LMSSA+CFF, we achieve the highest accuracy, which proves the effectiveness of our proposed modules. 

\begin{figure}[tb]\rmfamily
  \centering
  \includegraphics[width=\linewidth]{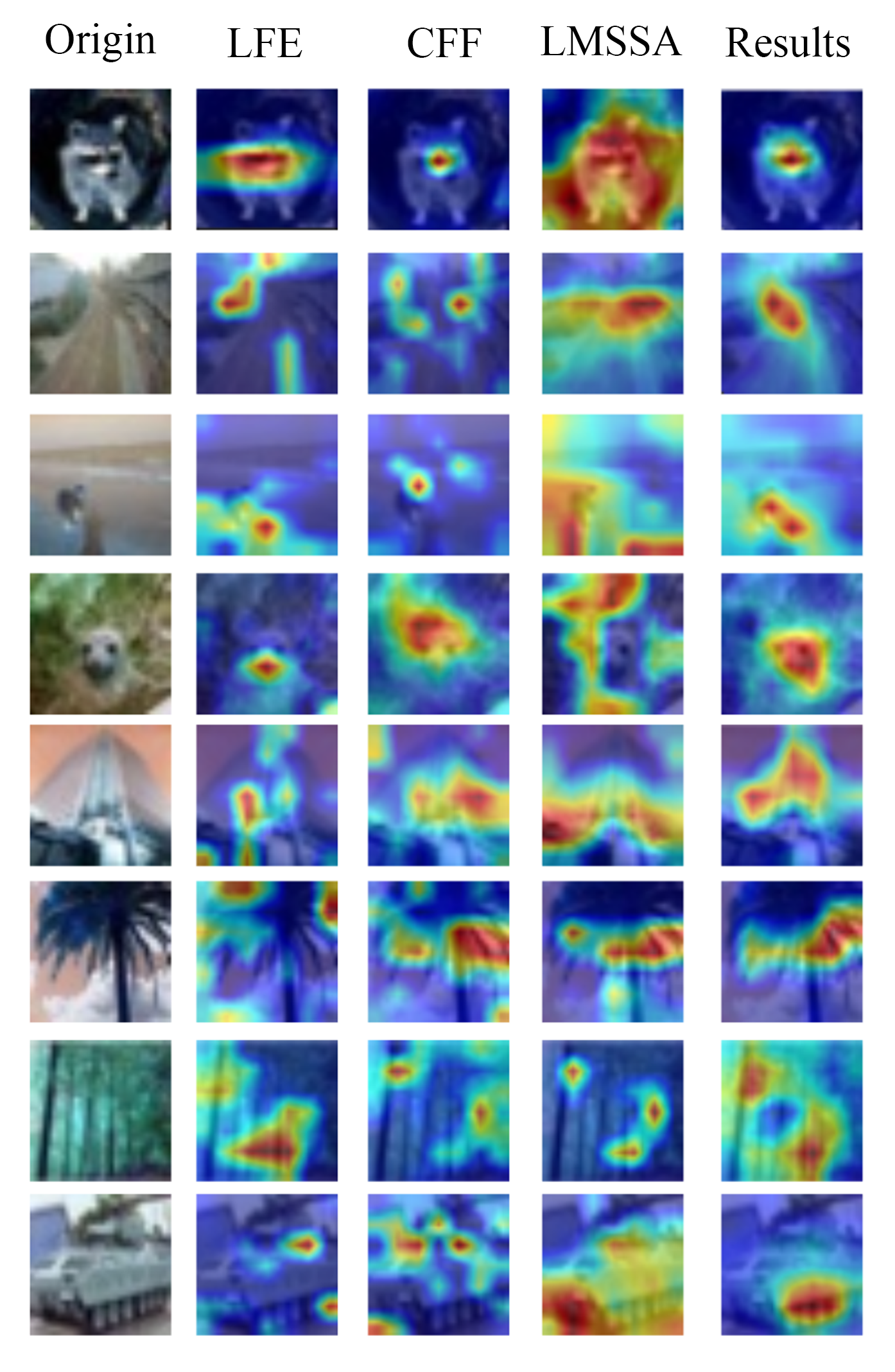}
  \caption{\rmfamily{Visual demonstration of the functions of the proposed LFE, CFF and LMSSA modules. }}
  \label{Fig6}
\end{figure}

\textcolor{red}{In order to visually demonstrate the function of the proposed LFE, LMSSA and CFF, a series of heatmaps has been drawn in Fig. \ref{Fig6} to display the attention results by using different components. We used a pre-trained model on CIFAR-100 and select 4 groups of images from the test set as out input (the $1^{\mathrm{st}}$ column), and use the output of the Stage 3 as the basic feature map. The $2^{\mathrm{nd}}$ to the $4^{\mathrm{th}}$ column show the results by using only LMSSA, CFF and LFE in Stage 4 respectively. The $5^{\mathrm{th}}$ column shows the final results by using them all. In particular, LFE reflects the features of the convolution layers. When the input contains a complex background, LMSSA pays more attention to the details. On the contrary, when the input image has a relatively simple background, it will pay more attention to the object. CFF pays more attention to the edge features of the object. }

\begin{table}[t]\rmfamily
\caption{Ablation experiment on the core components.}
\centering
\begin{tabular*}{\linewidth}{@{\extracolsep\fill}ccccc@{\extracolsep\fill}}
\hline
LFE	        & LMSSA	         & CFF            &GFLOPs&	Acc.            \\ \hline
            &                &                & 2.1  & 81.66            \\ 
            &   \checkmark   &                & 2.57 & 82.52            \\ 
 \checkmark &   \checkmark   &                & 2.58 & 82.77            \\ 
            &   \checkmark   &   \checkmark   & 2.49 & 83.23            \\
 \checkmark &   \checkmark   &   \checkmark   & 2.50 & \textbf{83.39}   \\ \hline
\end{tabular*}
\label{Tableab}
\end{table}

\subsubsection{The optimal kernel size of CNNs}\label{AB:CFF}
In this section, experiment has been conducted to explore the optimal size of convolutional kernels for efficient convolutional feature fusion with attention module. In the experiment, we remove LFE block to isolate the influence of convolution. Experiment has been conducted on CIFAR-100, following the settings of DeiT, we trained the models 200 epochs on a single NVIDIA RTX 3090 GPU.

The experimental results are shown in Table \ref{Table6}. The experimental results indicate that using 5×5 convolutional kernels in the first 2 layers of the network and 3×3 convolutional kernels in the following 2 layers yields better performance. The reason is that the 5×5 convolutional kernels have a larger receptive field, allowing them to extract more global information in the shallow layers, which can be better fused with the feature maps after attention calculation.

\begin{table}[tb]\rmfamily
\caption{The comparison of accuracy results by using different sizes of kernels.}
\centering
\begin{tabular*}{\linewidth}{@{\extracolsep\fill}ccccc@{\extracolsep\fill}}
\toprule
Kernel-size & 3 × 3 & 5 × 5 & 3×3/5×5 & 5×5/3×3 \\ \midrule
Acc & 82.77 & 82.46 & 82.69 & 83.23 \\ \bottomrule
\end{tabular*}
\label{Table6}
\end{table}

\subsubsection{Research on lightweight multi-scale self-attention}

We utilized deep convolutional layers for multi-scale self-attention feature fusion (instead of the original convolution approach). Additionally, we did not employ 1x1 convolutional kernels for channel restoration. To investigate the impact of these operations on accuracy on tiny datasets, we trained our model 200 epochs on CIFAR-100 following the setting outlined in section \ref{AB:CFF}. 

\begin{table}[tb]\rmfamily
\caption{ The performance comparison of different versions of models. }
\centering
\begin{tabular*}{\linewidth}{@{\extracolsep\fill}cccc@{\extracolsep\fill}}
\bottomrule
Type & Params (M) & GFLOPs & Acc \\ \midrule
lightweight & 14.0 & 2.50 & 83.39 \\ 
normal & 15.5(+10.7\%) & 2.63(+5.2\%) & 83.57(+0.18) \\ 
\bottomrule
\end{tabular*}
\label{Table7}
\end{table}

As shown in Table \ref{Table7}, using lightweight multi-scale self-attention leads to slight decrease in accuracy, but its reduces 10.7\% parameters and 5.2\% GFLOPs (such reduction will be even higher for larger sized models), i.e. our model strikes better trade-off between accuracy and model size. 

\subsubsection{The impact of the positional encoding}\label{POS}
To investigate the impact of PE during training on tiny datasets, we compare the performances of the models with or without PE. Besides, we remove CFF to prevent the potential impact of the inductive bias in convolutional structures (in CFF) on the experimental results of this section (the same reason for removing LFE in section \ref{AB:CFF}).  Then we analyze whether the local feature extraction block could replace the PE module. 

\begin{table}[tb]\rmfamily
\centering
\caption{The comparison of accuracy by using or without using PE.}
\begin{tabular*}{\linewidth}{@{\extracolsep\fill}ccc@{\extracolsep\fill}}
\toprule
Models & with PE & w/o PE \\ \midrule
Acc & 82.63 & 82.77 \\ \bottomrule
\end{tabular*}
\label{Table8}
\end{table}

As shown in Table \ref{Table8}, with the help of local feature extraction block, adding the PE module brings no obvious performance gain, indicating that the local feature extraction block could replace PE module in maintaining the accuracy. The reason is that the local feature extraction block implicitly contains the information present in PE.

It is noteworthy that in ablation study, all experiments were conducted on a single 3090 GPUs, therefore the corresponding results will be slightly lower than those in section \ref{Res}. Besides, we remove LFE in section \ref{AB:CFF} and remove CFF in section \ref{POS}, thus the results are consistent with those in Table \ref{Tableab} (without LFE or CFF). In section, we analyze the impact of lightweight self-attention mechanism on LMSSA, where neither LFE nor CFF are involved, hence the result (82.52) is consistent with the second row of Table \ref{Tableab}. 

\section{Conclusion}

A hybrid architecture (dubbed MSCViT) based on ViT is proposed in this paper, which aims to address the limitations of traditional Transformer models on tiny datasets. The proposed MSCViT leverages lightweight multi-scale attention along with convolutional fusion to capture both local and global information, thereby enhancing the network’s representational capacity. Compared with traditional ViT models and hybrid models, we achieve competitive results without stacking heavyweight modules. 

Extensive experimental results and comparisons with other popular models demonstrate the effectiveness and superiority of our proposed architecture. \textcolor{red}{It’s noteworthy that we aim at optimizing the model for tiny datasets, which may not be effective for large-scale datasets.} In the end, we hope our work paves the way for future research on tiny datasets.


\bibliographystyle{unsrt}
\bibliography{refs.bib}



\end{document}